%% file: anti_locality.tex
\documentclass[]{interact}

\pdfoutput=1 

\usepackage{epstopdf}
\usepackage[caption=false]{subfig}

\usepackage[natbibapa,nodoi]{apacite} 
\theoremstyle{plain}

\theoremstyle{definition}

\theoremstyle{remark}

\usepackage{tabularx}
\usepackage{xcolor}
\usepackage[latin1]{inputenc}
\usepackage{etoolbox} 

\begin{document}

\newtoggle{anonymous}
\togglefalse{anonymous}

\title{Anti dependency distance minimization in short sequences. A graph theoretic approach. }

\iftoggle{anonymous}{}
{ 
  \author{
  \name{Ramon Ferrer-i-Cancho\textsuperscript{a}\thanks{CONTACT Ramon Ferrer-i-Cancho. Email:   rferrericancho@cs.upc.edu} and Carlos G\'omez-Rodr\'iguez\textsuperscript{b}}
  \affil{\textsuperscript{a}Complexity and Quantitative Linguistics Lab, LARCA Research Group,  Departament de Ci\`encies de la Computaci\'o, Universitat Polit\`ecnica de Catalunya (UPC), Barcelona, Spain. ORCiD: 0000-0002-7820-923X; \textsuperscript{b}Universidade da Coru\~na, CITIC. FASTPARSE Lab, LyS Research Group, Departamento de Computaci\'on. Elvi\~na, 15071 A Coru\~na, Spain. ORCiD: 0000-0003-0752-8812}
  }
}
 
\maketitle

\begin{abstract}
Dependency distance minimization (DDm) is a word order principle favouring the placement of syntactically related words close to each other in sentences. Massive evidence of the principle has been reported for more than a decade with the help of syntactic dependency treebanks where long sentences abound. However, it has been predicted theoretically that the principle is more likely to be beaten in short sequences by the principle of surprisal minimization (predictability maximization). Here we introduce a simple binomial test to verify such a hypothesis. In short sentences, we find anti-DDm for some languages from different families. Our analysis of the syntactic dependency structures suggests that anti-DDm is produced by star trees.    
\end{abstract}

{\bf Keywords: dependency syntax; dependency distance minimization; word order; graph theory; treebanks}

\section{Introduction}

Dependency distance minimization (DDm) is a word order principle favouring the placement of syntactically related words close to each other in sentences \citep{Liu2017a}. Massive evidence of the principle has been reported for more than a decade with the help of syntactic dependency treebanks where long sentences abound \citep{Ferrer2004b,Liu2008a, Ferrer2013c,Futrell2015a}. Sometimes short sentences are excluded from the analyses (e.g., \cite{Jiang2015a}). See \cite{Liu2017a} for an overview of the cognitive origins of DDm. 

It has been argued theoretically that the principle would be easier to beat by other word order principles in at least two conditions: short sequences \citep{Ferrer2014a} and short words \citep{Ferrer2014e}, 
as dependency distances shorten and then the cognitive costs associated to them reduce, thus diminishing the pressure for DDm.
The aim of this article is to verify empirically the prediction that DDm should be beaten in short sequences. DDm is also known as dependency length minimization \citep{Futrell2015a}, but the term distance allows one to see DDm as particular case of a general principle of distance minimization, crucial for the construction of a parsimonious theory of language and cognition in general \citep{Ferrer2017c}.

A competitor of DDm is Sm, surprisal minimization, or PM, predictability maximization \citep{Levy2008a,Ferrer2013f}. Surprisal minimization is a less technical name for the principle of entropy minimization \citep{Ferrer2013f}.  Throughout this article we use {\em m} for minimization (as in Sm or DDm) and {\em M} for maximization (as in PM).  
A particular conflict between word order principles arises theoretically when deciding the placement of a head and its dependents. In single-head structures, the head should be put at the center according to the DDm principle whereas, according to the Sm or PM principles, the head should be put at one of the ends \citep{Ferrer2014a,Ferrer2013f}. For simplicity, Sm and PM are considered to be equivalent in this article but some subtle differences have been discussed theoretically \citep{Ferrer2013f}.

Focusing on the ordering of the verb (head) and its dependents (subject and object), such a conflict has been linked to the diversity of word orders, the existence of languages lacking a dominant word order, word order reversions in evolution, alternative word orders with the head at the center and the preference for head last in simple sequences and its loss in more complex sequences \citep{Ferrer2014a}. As for the latter, the rationale is that DDm would be more likely to win in long sequences leading to central head placements while Sm (or PM) would be more likely to win in very short sequences leading to non-central head placements.
A challenge for this argument is that DDm effects have also been found in short spans (such as noun phrases), casting doubts on the grounding of this effect in memory limitations \citep{Gulordava2015a}. Here we aim to verify the prediction that DDm is more likely to be beaten in short sequences with the help of real data with 75 languages from about 20 families. 

Anti-DDm has been investigated in depth in the cognitive science or psycholinguistics community under the umbrella of anti-locality effects \citep{Vasishth2006a, Rajkumara2016a}. However, such a research relies on psychological experiments suffering from a small set of languages and a limited range of set-ups or phenomena \citep{Liu2017a} in addition to considering only some but not all dependency distances \citep{Ferrer2017c}. Here we adopt a big data \citep{Liu2017a} and nomothetic \citep{Roberts2013b} approach where set-ups and languages are only limited by the growing collection of syntactic dependency treebanks employed.

Following our hypothesis, we focus on sentences of $n$ words with small $n$. $n = 3$ is a critical sentence length for a theory of word order because of DDm \citep{Ferrer2008e}. When $n < 2$ there is no word order problem at all. When $n = 2$, the distance between head and dependent is not affected by the order \citep{Ferrer2008e, Ferrer2014a}. Therefore, $n \geq 3$ is needed for the conflict between word orders above \citep{Ferrer2013f}.
Here we focus on the two smallest values of $n$ where such a conflict exists: $n=3$ and $n=4$. We exclude sentences of length from $n=5$ onwards for simplicity and also because DDm is more likely to manifest in longer sequences \citep{Ferrer2014a}. 

\begin{figure}
\begin{center}
\includegraphics[scale = 0.8]{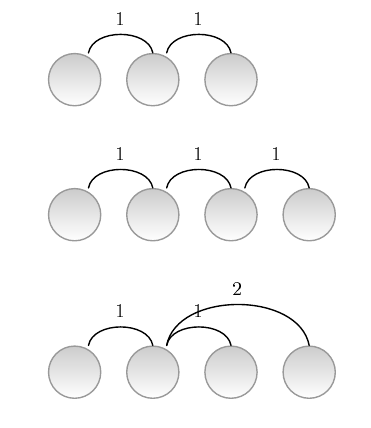}
\end{center}
\caption[Examples of trees with 3 and 4 vertices.]{\label{linear_and_star_tree_figure} Examples of trees with 3 and 4 vertices. Edge labels indicate edge distances. Top. A tree with 3 vertices that is both a linear tree and a star tree. Center: a linear tree with 4 vertices. Bottom: a star tree with 4 vertices.  }
\end{figure}

Although syntactic dependency structures are directed graphs \citep{melcuk88}, here we consider them as undirected for two reasons: dependency direction is not relevant in the calculation of dependency distances and it simplifies the analysis of the kinds of syntactic dependency structures.  When $n \in \{3,4\}$ the only structures that are possible are linear and star trees. 
A linear tree is a tree where the maximum degree is 2 whereas a star tree is a tree where the maximum degree is $n-1$, the maximum possible degree \citep{Ferrer2013d}. Fig. \ref{linear_and_star_tree_figure} shows some linear and star trees.  When $n=3$ the tree is both a linear tree and a star tree. Star trees correspond to the single-head structures where a conflict between DDm and Sm (or PM) has been demonstrated theoretically. The sum of dependency distances is maximized when the hub (the vertex of degree $n-1$) is put at one of the ends of the linear arrangement, which coincides with the arrangement minimizing surprisal (or maximizing predictability) \citep{Ferrer2014a,Ferrer2013f}.

If the theoretical arguments above are correct, one would expect to find dependencies that are farther than expected by chance when $n \in \{3,4\}$ for two reasons. First, the abundance of star trees on which the theoretical conflict above holds \citep{Ferrer2014a}. Second, the smaller dependency distances that are expected as a side effect of the short length of the sentences \citep{Ferrer2014a}. In its current state of development, that theory has no specific predictions to make on linear trees.

We define chance with respect to some null models. Our core null model is a random linear arrangement of the vertices where the syntactic dependency structure of sentences remains constant \citep{Ferrer2004b, Ferrer2018a}. When $n=3$, the syntactic dependency structure is always the same (a tree that is both a linear tree and a star tree). As this does not happen when $n = 4$, we also consider an additional null model for $n = 4$ where not only the order but also the syntactic dependency structure (a star tree or a linear tree) is chosen at random.   

When trying to shed light on the origins of anti-DDm in languages, various complementary approaches are possible, e.g., psychological experiments or traditional linguistic analyses based on the properties of the vertices (e.g., their part-of-speech) or the type of the dependencies. Here we adopt a graph theoretic approach that abstracts away from these properties to allow one to maximize the generality and parsimony of potential explanations. Indeed, we will show that the kind of graph structure apparently determines the possibility that anti-DDm emerges in short sequences.
Our graph theoretic approach is radical in the sense that we focus on aspects of the graph structure that can be analyzed independently from the linear ordering of the vertices.
Common features in research on dependency syntax such as dependency distance, branching direction or adjacency (e.g., \cite{Jiang2015a}) depend on that ordering. Examples of features that do not depend on it are hubiness \citep{Ferrer2013b,Ferrer2017a}, hierarchical distance \citep{Jing2015a} or whether the syntactic dependency structure is a star tree or a linear tree (in this article).

When investigating DDm, many researchers have considered a stronger null model where, for instance, dependency crossings are not allowed. A popular example of this tradition is the recent work by \citet{Futrell2015a}. However, we have argued that this could shadow the very effects of DDm \citep{Ferrer2016a}. It is crucial for our study to use a null model that does not introduce a bias for or against DDm. Thus, the only constraint of our null model for a given sentence is that all $n!$ possible orderings are equally likely, as in the pioneering research of one of us \citep{Ferrer2004b}, and as expected from the maximum entropy principle without constraints \citep{Kesavan2009a}.

The remainder of the article is organized as follows. Section \ref{materials_section} presents the syntactic dependency treebanks, i.e. collections of sentences with syntactic dependency annotations, that we used to investigate biases against DDm. By the conflict above, evidence of anti-DDM can be interpreted as Sm (or PM) beating DDm in star trees at least. Therefore, when we use the term anti-DDm we are not referring to a cognitive principle or a principle of word order but rather to a statistical phenomenon that can be attributed to the Sm (or PM) principle.

Section \ref{methods_section} presents the statistical methods used, introducing a new binomial test that allows one to detect biases against or for DDm. 
To detect anti-DDm, the test examines the number of sentences where the sum of dependency distances is above the expected value in a random linear arrangement. If that number is significantly large, the test concludes that there is evidence of anti-DDm. Similarly, to detect DDm, the test examines the number of sentences where the sum of dependency distances is below the expected value in a random linear arrangement. If that number is significantly large, the test concludes that there is evidence of DDm.
Section \ref{methods_section} shows that if dependency crossings were not allowed, the tests would lose statistical power (they would become more conservative) when $n=4$ (crossings for $n \le 3$ are impossible \citep{Ferrer2013b}), shadowing either the effects of DDm or the effects of biases against DDm, supporting previous arguments for the case of DDm \citep{Ferrer2016a}.
Section \ref{results_section} shows that some languages from different families exhibit an anti-DDm effect even after controlling for multiple comparisons. Interestingly, we find anti-locality in languages for which anti-locality effects have never been reported before based on traditional psychological experiments. Furthermore, our analysis of trees of four vertices suggests that anti-DDm is produced by star trees. 

\section{Data}

\label{materials_section}

\input{materials}

\section{Methods}

\label{methods_section}

\input{methods}

\section{Results}

\label{results_section}

\begin{table}
\caption[Binomial test results rejecting the null hypothesis in favour of anti-DDm.]{\label{binomial_test_above_table} For every collection and every level of analysis, we show $l_0$, the number of languages, $l$, the number of languages where the number of sentences available reaches $s^*$, $f$, the number of languages where the binomial test rejects the null hypothesis in favour of anti-DDm at a significance level of 0.05, $f_H$, the same as $f$ after applying the Holm correction to the languages counted in $f$. 
The number attached to the language name indicates the magnitude of the corrected p-value. It is obtained after rounding $-log_{10}(\mbox{p-value})$ to leave just one decimal digit. Then the significance level $\alpha = 0.05$ gives 1.3.  
}
{\scriptsize
\begin{tabularx}{\linewidth}{llllllllX}
Collection & $n$ & Kind & $l_0$ & $l$ & $f$ & $f_H$ & Family & Languages\\
\hline
UD & 3 & all & 73 & 64 & 9 & 6 & Austronesian (1) & Tagalog$^{1.5}$ \\
 & & & & & & & Dravidian (1) & Telugu$^{1.8}$ \\
 & & & & & & & Indo-European (3) & English$^{4.6}$ Old French$^{7.2}$ Slovak$^{6.2}$ \\
 & & & & & & & Japanese (1) & Japanese$^{42.9}$ \\
 & 4 & all & 75 & 70 & 1 & 0 & --- & --- \\
 & 4 & unlabelled & 75 & 70 & 1 & 1 & Pama-Nyungan (1) & Warlpiri$^{2.4}$ \\
 & 4 & labelled & 75 & 71 & 10 & 2 & Indo-European (1) & English$^{1.9}$ \\
 & & & & & & & Pama-Nyungan (1) & Warlpiri$^{3.4}$ \\
 & 4 & star & 72 & 64 & 19 & 11 & Dravidian (1) & Telugu$^{7}$ \\
 & & & & & & & Indo-European (9) & Bulgarian$^{5.1}$ Czech$^{3.8}$ English$^{4.2}$ Faroese$^{2}$ French$^{5.5}$ Old French$^{17}$ Portuguese$^{1.8}$ Russian$^{2}$ Slovak$^{5.7}$ \\
 & & & & & & & Mande (1) & Bambara$^{1.3}$ \\
 & 4 & linear & 75 & 68 & 0 & 0 & --- & --- \\
Stanford & 3 & all & 30 & 29 & 7 & 5 & Dravidian (1) & Telugu$^{8.2}$ \\
 & & & & & & & Indo-European (3) & Czech$^{4.2}$ German$^{5.2}$ Slovak$^{12}$ \\
 & & & & & & & Japanese (1) & Japanese$^{25.5}$ \\
 & 4 & all & 30 & 29 & 1 & 0 & --- & --- \\
 & 4 & unlabelled & 30 & 29 & 1 & 0 & --- & --- \\
 & 4 & labelled & 30 & 30 & 5 & 3 & Dravidian (1) & Telugu$^{2.5}$ \\
 & & & & & & & Indo-European (2) & English$^{2.1}$ Ancient Greek$^{8.5}$ \\
 & 4 & star & 30 & 29 & 11 & 9 & Dravidian (1) & Telugu$^{11.2}$ \\
 & & & & & & & Indo-European (8) & Bengali$^{3}$ Bulgarian$^{1.5}$ Czech$^{3}$ English$^{3.7}$ Ancient Greek$^{3.2}$ Latin$^{1.8}$ Portuguese$^{2.7}$ Slovak$^{16.1}$ \\
 & 4 & linear & 30 & 29 & 0 & 0 & --- & --- \\
Prague & 3 & all & 30 & 29 & 3 & 2 & Dravidian (1) & Telugu$^{13.8}$ \\
 & & & & & & & Indo-European (1) & Persian$^{14.7}$ \\
 & 4 & all & 30 & 29 & 3 & 2 & Dravidian (1) & Telugu$^{2.7}$ \\
 & & & & & & & Indo-European (1) & Ancient Greek$^{2}$ \\
 & 4 & unlabelled & 30 & 29 & 2 & 2 & Dravidian (1) & Telugu$^{4}$ \\
 & & & & & & & Indo-European (1) & Ancient Greek$^{2.4}$ \\
 & 4 & labelled & 30 & 30 & 3 & 2 & Dravidian (1) & Telugu$^{9.8}$ \\
 & & & & & & & Indo-European (1) & Ancient Greek$^{10.1}$ \\
 & 4 & star & 30 & 28 & 5 & 4 & Dravidian (1) & Telugu$^{22.8}$ \\
 & & & & & & & Indo-European (3) & Bengali$^{2.3}$ Ancient Greek$^{6.4}$ Persian$^{16.4}$ \\
 & 4 & linear & 30 & 29 & 0 & 0 & --- & --- \\
\end{tabularx}
}
\end{table}

Table \ref{binomial_test_above_table} summarizes the analysis of anti-DDm within each collection showing the number of languages for where the binomial test rejects the null hypothesis at a significance level of 0.05. 
The null hypothesis is impossible to reject due to undersampling only in a few languages (the difference between $l_0$ and $l$ is small, if any).
After controlling for multiple comparisons, anti-DDm is found in some languages (as indicated by the value of $f_H$ in Table \ref{binomial_test_above_table}). 
For instance, at the level of $n=3$ with UD dependencies, anti-DDm is found in $f=9$ languages but after controlling for multiple comparisons it survives in $f_H=6$ languages. 
When $n=4$, it turns out that anti-DDm is never found in linear trees but found in star trees. Star trees with $n=4$ are the level of analysis with the highest support for anti-DDm ($f_H$ is the highest in each collection according to Table \ref{binomial_test_above_table}).

Two findings reduce the chance that the results are due to a common descent \citep{Roberts2013b}. 
First, the languages where anti-DDm is found belong to different families. 
Second, the anti-locality does not cover a whole family if the family is represented by more than one language {\em a priori} according to Table \ref{typological_diversity_table}. 
For instance, at the level of $n=3$ with UD dependencies, the $f_H=6$ languages with evidence of anti-DDm are three Indo-European languages out of 44 (English, French and Slovak), one Austronesian language out of 2 (Tagalog), one Dravidian language out of 2 (Telugu) and one isolate (Japanese).

Table \ref{binomial_test_above_table} confirms the prediction that anti-DDm should surface more clearly when the uniformly random trees are labelled trees instead of unlabelled trees (Figure \ref{labelled_unlabelled_figure}) are used for reference. At the level of $n=4$ with UD dependencies, anti-DDm is found in $f=10$ languages before  controlling for multiple comparisons and $f_H=2$ languages after controlling for that when random labelled trees are used for reference. In contrast, anti-DDm is found in only one language before and after controlling for multiple comparisons when random unlabelled trees are used for reference. 

Given these findings, we aim to evaluate the scope of DDm using the same kind of binomial tests. Table \ref{binomial_test_below_table} shows a broader support for DDm. The weakest support is found at the levels $n=3$ and star trees with $n=4$ according to the value of $f_H$ in Table \ref{binomial_test_below_table}. Indeed, Table \ref{binomial_test_below_table} shows an opposite behavior with respect to Table \ref{binomial_test_above_table}: when $n=4$, support for DDm is stronger in linear trees than star trees ($f_H$ is greater in linear trees).  
  
\begin{table}
\caption[Binomial test results rejecting the null hypothesis in favour of DDm.]{\label{binomial_test_below_table} For every collection and every level of analysis, we show $l_0$, the number of languages, $l$, the number of languages where the number of sentences available reaches $s^*$, $f$, the number of languages where the binomial test rejects the null hypothesis in favour of DDm at a significance level of 0.05, $f_H$, the same as $f$ after applying the Holm correction to the languages counted in $f$. The number attached to the language family indicates the number of languages of the family with a corrected p-value below the significance level. Full names of languages are not shown due to pressure of space.
}
{\scriptsize
\begin{tabularx}{\linewidth}{lllllllX}
Collection & $n$ & Kind & $l_0$ & $l$ & $f$ & $f_H$ & Families \\
\hline
UD & 3 & all & 73 & 69 & 28 & 16 & Afro-Asiatic (3) Altaic (1) Basque (1) Indo-European (6) Mande (1) Sino-Tibetan (1) Uralic (3)  \\
 & 4 & all & 75 & 70 & 60 & 50 & Afro-Asiatic (4) Altaic (3) Austro-Asiatic (1) Austronesian (1) Basque (1) Indo-European (29) Japanese (1) Korean (1) Mongolic (1) Other (1) Sino-Tibetan (2) Uralic (5)  \\
 & 4 & unlabelled & 75 & 70 & 59 & 49 & Afro-Asiatic (4) Altaic (3) Austro-Asiatic (1) Austronesian (1) Basque (1) Indo-European (28) Japanese (1) Korean (1) Mongolic (1) Other (1) Sino-Tibetan (2) Uralic (5)  \\
 & 4 & labelled & 75 & 71 & 61 & 59 & Afro-Asiatic (4) Altaic (3) Austro-Asiatic (1) Austronesian (1) Basque (1) Dravidian (1) Indo-European (35) Japanese (1) Korean (1) Mande (1) Mongolic (1) Other (1) Sign Language (1) Sino-Tibetan (2) Uralic (5)  \\
 & 4 & star & 72 & 64 & 21 & 15 & Afro-Asiatic (2) Altaic (1) Austro-Asiatic (1) Basque (1) Indo-European (5) Sino-Tibetan (2) Uralic (3)  \\
 & 4 & linear & 75 & 68 & 61 & 54 & Afro-Asiatic (4) Altaic (3) Austro-Asiatic (1) Austronesian (1) Basque (1) Dravidian (1) Indo-European (31) Japanese (1) Korean (1) Mande (1) Mongolic (1) Other (1) Sino-Tibetan (2) Uralic (5)  \\
Stanford & 3 & all & 30 & 30 & 8 & 6 & Altaic (1) Basque (1) Indo-European (3) Uralic (1)  \\
 & 4 & all & 30 & 29 & 24 & 24 & Afro-Asiatic (1) Altaic (1) Basque (1) Indo-European (17) Japanese (1) Uralic (3)  \\
 & 4 & unlabelled & 30 & 29 & 24 & 24 & Afro-Asiatic (1) Altaic (1) Basque (1) Indo-European (17) Japanese (1) Uralic (3)  \\
 & 4 & labelled & 30 & 30 & 29 & 28 & Afro-Asiatic (1) Altaic (1) Basque (1) Dravidian (1) Indo-European (20) Japanese (1) Uralic (3)  \\
 & 4 & star & 30 & 29 & 8 & 7 & Afro-Asiatic (1) Basque (1) Indo-European (4) Uralic (1)  \\
 & 4 & linear & 30 & 29 & 28 & 28 & Afro-Asiatic (1) Altaic (1) Basque (1) Dravidian (1) Indo-European (20) Japanese (1) Uralic (3)  \\
Prague & 3 & all & 30 & 30 & 20 & 19 & Afro-Asiatic (1) Altaic (1) Basque (1) Indo-European (13) Japanese (1) Uralic (2)  \\
 & 4 & all & 30 & 29 & 23 & 23 & Afro-Asiatic (1) Altaic (1) Basque (1) Indo-European (16) Japanese (1) Uralic (3)  \\
 & 4 & unlabelled & 30 & 29 & 24 & 23 & Afro-Asiatic (1) Altaic (1) Basque (1) Indo-European (16) Japanese (1) Uralic (3)  \\
 & 4 & labelled & 30 & 30 & 26 & 25 & Afro-Asiatic (1) Altaic (1) Basque (1) Indo-European (18) Japanese (1) Uralic (3)  \\
 & 4 & star & 30 & 28 & 15 & 11 & Basque (1) Indo-European (8) Uralic (2)  \\
 & 4 & linear & 30 & 29 & 28 & 28 & Afro-Asiatic (1) Altaic (1) Basque (1) Dravidian (1) Indo-European (20) Japanese (1) Uralic (3)  \\
\end{tabularx}
}
\end{table}



\section{Discussion}

\label{discussion_section}

We have confirmed the theoretical prediction that anti-DDm should be found in short sequences \citep{Ferrer2014a}: we have found  evidence of anti-DDm in short sequences in languages from different families suggesting that anti-DDm is not lineage specific. The fact that anti-DDm is found in all annotation formalisms and that some languages show anti-DDm for more than one formalism (Telugu shows anti-DDm in all formalisms; in addition Ancient Greek shows anti-DDm in both Prague and Stanford dependencies) suggests that differences in annotation criteria cannot explain exclusively our findings.  
Interestingly, we have found anti-DDm in Telugu, Tagalog, French, Slovak in UD for $n=3$ (Table \ref{binomial_test_above_table}) that are languages for which anti-locality effects have never been reported before based on traditional psychological experiments as far as we know. 
Similar arguments can be made for $n = 4$. These discoveries illustrate the power of our statistical approach. 

Following a classic view of memory, it could be argued that DDm is not activated in short sentences because they do not exhaust the capacity of short-term memory. Sentences of length 3 and 4 have a number of words that fits into the magical number 4 in short-term memory \citep{Cowan2001a}. That memory limit has been has been argued to have been confirmed by the fact that mean dependency distance is below 4 \citep{Jing2015a} but such a limit could be directly related to the breakpoint in the decay of the probability of dependency distance that is found at distance 4-5 \citep{Ferrer2004b,Ferrer2017d}. Tentatively, if DDm were the only word order principle, one would expect an arbitrary ordering of words (a random ordering), but it has been shown that order in short sequences is lawful (e.g. \cite{Goldin-Meadow2008a,Langus2010a}). Such a lawfulness is confirmed by our finding of DDm and, to a lower degree, of anti-DDm. A reason for finding DDm even in short sentences could be some version of the Performance-Grammar Correspondence Hypothesis (PGCH): {\em ``grammars have conventionalized syntactic structures in proportion to their degree of preference in performance, as evidenced by patterns of selection in corpora and by ease of processing in psycholinguistic experiments''} \citep[3]{Hawkins2004a}. From an evolutionary standpoint, it could be that languages have undergone general adaptions consistent with DDm even in short sentences. But then, why should there be anti-DDm in short sequences as we have found? 
The conflict between Sm and DDm could explain it (short-term memory and DDm alone cannot): if short-term memory is not a problem any more, if would be easier for Sm to surface leading to a placement of the heads at one of the ends of the sequence \citep{Ferrer2013f}. It has been argued theoretically that pressure for DDm should be smaller in short sequences \citep{Ferrer2014a}, which could explain why the conflict between Sm and DDm \citep{Ferrer2013f} is resolved in favour of DDm in short sentences (in the absence of general adaptations for DDm across all scales). Similar arguments can be used to shed light on experiments on unconventional gestural communication with short sequences where the head tends to be put at the end, against DDm \citep{Ferrer2013f}.

Our investigation of the effect of tree structure, linear tree versus star trees, is a further step into understanding lawfulness in short sequences. Indeed, our
analysis of the case $n=4$ clarifies the nature of anti-DDm: anti-DDm is never found in linear trees but found in star trees (Table \ref{binomial_test_above_table}) whereas DDm is much stronger on linear trees than star trees (Table \ref{binomial_test_below_table}). This provides indirect empirical support for the theoretical conflict between DDm and Sm in the simple setup where it was proposed: one head and $n-1$ dependents, which implies star trees. When DDm wins the head should be put at the center; when Sm over heads wins, the head should be put last; when Sm over dependents wins, the head should be put first \citep{Ferrer2013f}. 
We hypothesize that anti-DDm is never found in linear trees with $n=4$ because the conflict between Sm and DDm reduces or disappears completely, namely the optima of DDm and Sm are closer or even coincide for linear trees.
The fact that a star tree does not imply a single head with $n-1$ dependents (the root of the syntactic dependency structure may not be the hub, the most connected node) suggests that directed syntactic dependency structures should be the subject of future research. Here we have chosen undirected structures for simplicity as the first step of a new research line. 

Another reason to not find anti-DDm in linear trees is that DDm may not be acting only at the level of the ordering of the words of the sentence but also at the level of the tree structures. The fact that $D_{min}$, the minimum value of $D$ for a given tree, is minimized by linear trees \citep{Ferrer2013b}, where 
\begin{equation*}
D_{min} = n-1,
\end{equation*}
and maximized by star trees \citep{Esteban2016a}, where 
\begin{equation*}
D_{min} = \frac{n^2 - n \bmod 2}{4},
\end{equation*}
suggests that DDm could be favouring the choice of linear trees to ease the optimization problem.  

It has been argued that nomothetic studies (statistical analyses of large-scale, cross-cultural data) like ours should be seen as hypothesis generating tools rather than as standalone studies due to the inter-connectedness of cultural traits \citep{Roberts2013b}.
Ideally, hypotheses should be generated from theory \citep{Roberts2013b}. In our case, we have used cross-linguistic data to test a prior theoretical hypothesis on the competition between word order principles in short sequences. Based on our analysis of the origins of the findings, we would like to invite researchers to confirm by means of lab experiments that anti-locality effects are practically missing in linear trees while found in star trees.  
 
Here we have investigated anti-DDm in short sentences. The same methodology could be applied to phrases or constituents that are also short \citep{Gulordava2015a}. This could help to find anti-DDm in more languages. 

\iftoggle{anonymous}{
}
{
\section*{Acknowledgements}

We are very grateful to G. J\"ager for his hospitality and rich discussions from many perspectives. We also thank D. Celinska-Kopczynska for helpful discussions on the problem of multiple comparisons and many suggestions to improve the article. The manuscript has benefited enormously from the comments of an anonymous reviewer.
RFC is supported by the grant TIN2017-89244-R from MINECO (Ministerio de Economia, Industria y Competitividad) and the recognition 2017SGR-856 (MACDA) from AGAUR (Generalitat de Catalunya).
CGR has received funding from the European Research Council (ERC), under the European Union's Horizon 2020 research and innovation
programme (FASTPARSE, grant agreement No 714150), from the ANSWER-ASAP project (TIN2017-85160-C2-1-R) from MINECO, and from Xunta de Galicia
(ED431B 2017/01, and a grant from Conseller\'ia de Cultura, Educaci\'on e Ordenaci\'on Universitaria to complement ERC grants).
}

\bibliographystyle{apacite}
\bibliography{../biblio_dt/main,../biblio_dt/twoplanaracl,../biblio_dt/Ramon}


\end{document}

%% file: materials.tex
In order to provide results on a wide range of languages of various families, while also controlling for the possible effects of differences in syntactic annotation criteria, we analyze two different collections of treebanks:
\begin{itemize}
\item Universal Dependencies (UD) 2.3 \citep{UD23truncated}, the largest and most diverse dependency treebank collection that is currently available. It is comprised of 129 treebanks of 76 languages, annotated following the Universal Dependencies guidelines.
\item HamleDT 2.0 \citep{HamledTStanford}, a collection of treebanks from 30 languages, each of them annotated with two different sets of guidelines: Universal Stanford dependencies \citep{UniversalStanford} and Prague dependencies \citep{PDT20}. In tables, we will simply write \emph{Prague} and \emph{Stanford} to refer to the HamleDT collection with Prague and Stanford annotation, respectively.
\end{itemize}
Universal Stanford dependencies are closely related to UD, as the latter evolved out of the former, which in turn are a multilingual adaptation of the Stanford Dependencies for English \citep{Stanford2008}, based on lexical-functional grammar \citep{Bresnan00}. However, Prague dependencies provide significantly different structural representations, based on the functional generative description \citep{Sgall69} of the Praguian linguistic tradition \citep{Hajicova95}. In terms of tree structure, the most relevant differences are the annotation of conjunctions and adpositions \citep{HowFarStanfordPrague}.

Note that for many languages, there are UD and HamleDT treebanks with overlapping source material. Thus, our main goal in including HamleDT 2.0 is to provide results with different annotation formalisms, rather than to provide more data or languages with respect to using only UD. In this respect, it is also worth noting that, while a more recent version of HamleDT exists (3.0), it abandoned the dual annotation and adopted Universal Dependencies (version 1.1) as its only annotation style, so this newer version is not useful for our purposes. 

For our analysis, punctuation tokens are removed from the treebanks, following common practice in research on statistical properties of dependency structures \citep{Gomez2016a}. Nodes that do not represent words, such as the null elements present in the Bengali, Hindi and Telugu HamleDT corpora and the empty nodes in various Universal Dependencies treebanks, are also removed. To preserve the integrity of dependency structures, non-deleted nodes whose head has been deleted are reattached as dependents of their nearest non-deleted ancestor.

Table \ref{typological_diversity_table} summarizes the linguistic diversity of our collections of treebanks. Bengali is the only language in the HamleDT collection that is not present in UD.
 
The UD collection contains sentences in 76 languages, belonging to 18 families. 
However, we exclude Yoruba from our analysis because its treebank does not contain any sentences of length 3 or 4. The remaining 75 languages belong to 17 families.  
Among these languages, there are a few special cases. One is Naija, an English-based pidgin language spoken in Nigeria, to which we assign the family \emph{Other}. There is also a treebank of the Swedish Sign Language, which we associate with a family {\em Sign Language} for being the only non-vocal language in the sample. Finally, one of the corpora corresponds to a code-switching variety (Hindi-English), which belongs to the Indo-European family that is common to both languages.

The Prague and the Stanford HamleDT collections have 30 languages from 7 families. 

Tables \ref{binomial_test_above_table} and \ref{binomial_test_below_table} show that the original number of languages (75) reduces to some number $l_0$ depending on the level of analysis in UD. When $n=3$, the number of languages drops from 75 to 73 (there are two languages lacking sentences of length 3). When $n = 4$, the number of languages remains unchanged at the level of linear trees ($l_0 = 75$) while it drops from 75 to 72 at the level of star trees (there are three languages lacking star trees of 4 vertices).
In the other collections, $l_0$ matches the original number (30) in all cases. 

\begin{table}
\caption[Typological diversity of the treebank collections used in this study.]{\label{typological_diversity_table}The languages in every collection sorted by family. The counts attached to the collection names indicate the number of different families and the number of different languages. The counts attached to family names indicate the number of different languages. }
{\scriptsize 
\begin{tabularx}{\linewidth}{llX}
Collection & Family & Languages \\
\hline
UD (18, 76) & Afro-Asiatic (6) & Akkadian Amharic Arabic Coptic Hebrew Maltese \\
  & Altaic (3) & Kazakh Turkish Uyghur \\
  & Austro-Asiatic (1) & Vietnamese \\
  & Austronesian (2) & Indonesian Tagalog \\
  & Basque (1) & Basque \\
  & Dravidian (2) & Tamil Telugu \\
  & Indo-European (44) & Afrikaans Ancient Greek Armenian Belarusian Breton Bulgarian Catalan Croatian Czech Danish Dutch English Faroese French Galician German Gothic Greek Hindi Hindi-English Irish Italian Kurmanji Latin Latvian Lithuanian Marathi Norwegian Old Church Slavonic Old French Persian Polish Portuguese Romanian Russian Sanskrit Serbian Slovak Slovenian Spanish Swedish Ukrainian Upper Sorbian Urdu \\
  & Japanese (1) & Japanese \\
  & Korean (1) & Korean \\
  & Mande (1) & Bambara \\
  & Mongolic (1) & Buryat \\
  & Niger-Congo (1) & Yoruba \\
  & Other (1) & Naija \\
  & Pama-Nyungan (1) & Warlpiri \\
  & Sign Language (1) & Swedish Sign Language \\
  & Sino-Tibetan (2) & Cantonese Chinese \\
  & Tai-Kadai (1) & Thai \\
  & Uralic (6) & Erzya Estonian Finnish Hungarian Komi Zyrian North Sami \\
Stanford (7, 30) & Afro-Asiatic (1) & Arabic \\
  & Altaic (1) & Turkish \\
  & Basque (1) & Basque \\
  & Dravidian (2) & Tamil Telugu \\
  & Indo-European (21) & Ancient Greek Bengali Bulgarian Catalan Czech Danish Dutch English German Greek Hindi Italian Latin Persian Portuguese Romanian Russian Slovak Slovenian Spanish Swedish \\
  & Japanese (1) & Japanese \\
  & Uralic (3) & Estonian Finnish Hungarian \\
Prague (7, 30) & Afro-Asiatic (1) & Arabic \\
  & Altaic (1) & Turkish \\
  & Basque (1) & Basque \\
  & Dravidian (2) & Tamil Telugu \\
  & Indo-European (21) & Ancient Greek Bengali Bulgarian Catalan Czech Danish Dutch English German Greek Hindi Italian Latin Persian Portuguese Romanian Russian Slovak Slovenian Spanish Swedish \\
  & Japanese (1) & Japanese \\
  & Uralic (3) & Estonian Finnish Hungarian \\
\end{tabularx}
}
\end{table}

%% file: methods.tex
\subsection{Graph theory}

\label{graph_theory_subsection}

\begin{figure}
\begin{center}
\includegraphics[scale = 0.8]{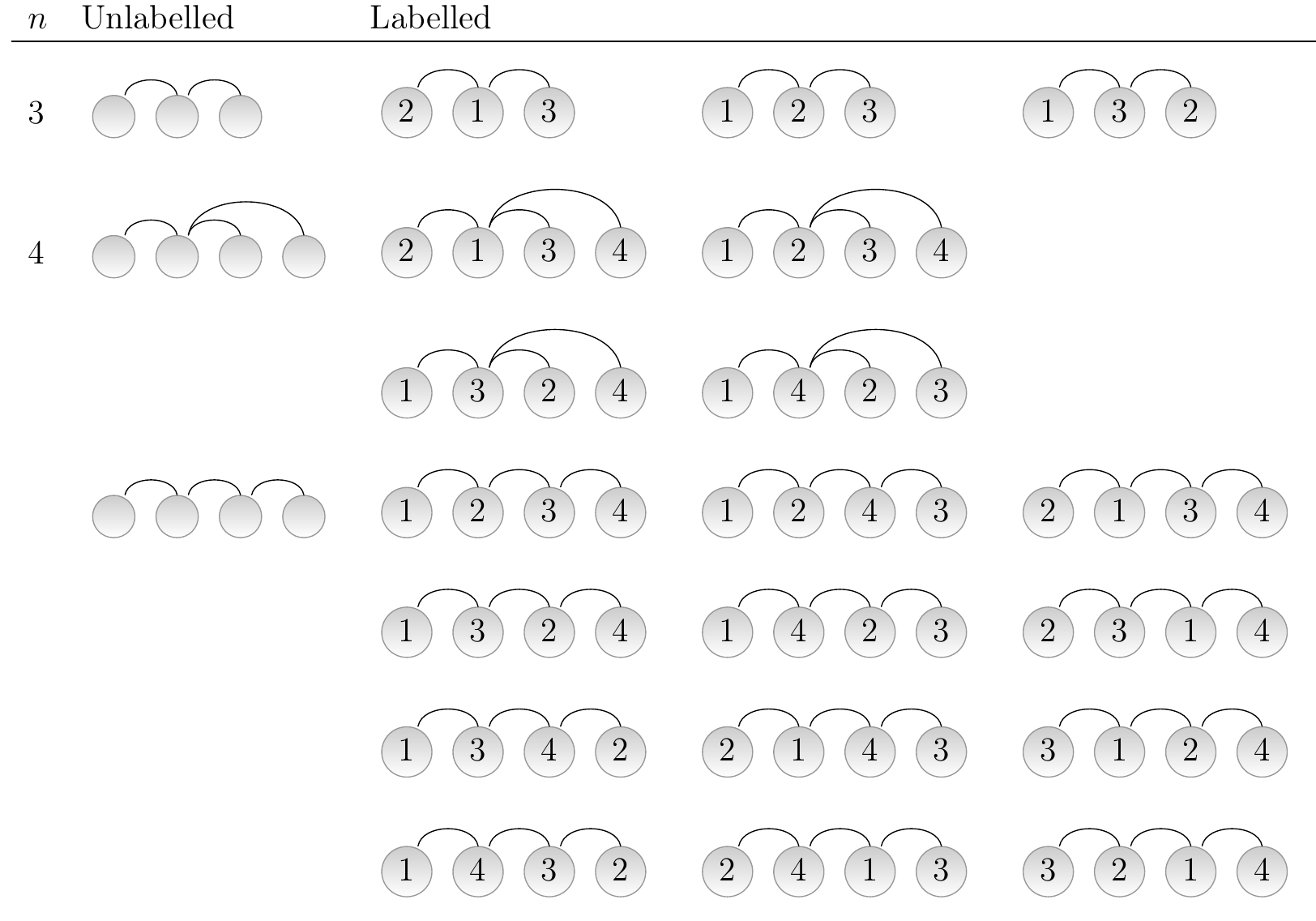}
\end{center}
\caption[All the distinct labelled and unlabelled trees of 3 and 4 vertices.]{\label{labelled_unlabelled_figure} All the distinct labelled and unlabelled trees of 3 and 4 vertices. The enumeration for $n=4$ is adapted from \cite{Longani2008a}.}
\end{figure}

A leaf is a vertex of degree 1 and an internal vertex is a vertex of degree greater than 1. We use the term hub to refer to the vertex of a tree having the largest degree \citep{Ferrer2013b}. In a star tree, the are $n - 1$ leaves and the only internal vertex is the hub.

When deciding when two trees are the same, standard theory provides two well-known criteria. One is when the vertices of a tree are labelled with $n$ distinct numbers, playing the role of vertex identifiers. In this case, the trees are said to be labelled. The other criterion is when the vertices are unlabelled (they have no identifier). Figure \ref{labelled_unlabelled_figure} shows the distinct trees for each of the two criteria. When $n=3$ there is only one possible unlabelled tree but three labelled trees. Each distinct labelled tree is defined by the label assigned to the internal vertex. By symmetry, exchanging the identifiers of the leaves does not produce another labelled tree. A distinct tree is only produced when the labels of a leaf and an internal vertex are exchanged.
When $n = 4$, there are two unlabelled trees and 16 labelled trees. The unlabelled trees are a star tree and a linear tree. The unlabelled star tree yields 4 different labelled star trees that are 	determined by the label of the hub. As before, exchanging the labels of the leaves does not produce a new labelled tree. A distinct tree is only obtained swapping the label of the hub with that of another vertex. The unlabelled linear tree yields 12 different labelled linear trees that are determined by all the permutations of the four labels and the fact that each permutation and its reverse correspond to the same labelled linear tree, therefore there are $4!/2 = 12$ distinct labelled linear trees.

Unlabelled trees are the most abstract criterion to define what a distinct tree is. Labelled trees define a level of abstraction that is intermediate between that of unlabelled trees and the linear arrangement of vertices. Notice that the labels of a labelled tree can be interpreted as vertex positions and thus define a linear arrangement of the vertices. However, the labellings in all the labelled trees of a given unlabelled tree do not cover all possible linear arrangements. For instance, Figure \ref{labelled_unlabelled_figure} shows that a star tree of 4 vertices has 4 different labelled trees but there are actually $4!=24$ possible linear arrangements of a star tree. In particular, each labelled star tree of 4 vertices can be arranged linearly in $3!$ different ways once the hub is placed in the position defined by its label.

In the context of trees whose vertices are assigned distinct positions in a linear arrangement (thus defining a particular labelled tree where labels indicate vertex positions),
we define the distance of an edge as the distance in vertices between the linked vertices forming the edge: consecutive vertices are at distance 1, vertices separated by a vertex are at distance 2 and so on \citep{Ferrer2004b}. 
Suppose that $D$ is the sum of edge distances of a syntactic dependency tree \citep{Ferrer2014f}, i.e.
\begin{equation*}
D = \sum_{i=1}^{n-1} d_i,
\end{equation*}
where $d_i$ is the distance of the $i$-th edge. In Fig. \ref{linear_and_star_tree_figure}, the tree of 3 vertices has $D = 1 + 1 = 2$, the linear tree of 4 vertices has $D=1+1+1 = 3$ and the star tree of 4 vertices has $D= 1+1+2 = 4$.

There are $n!$ linear arrangements of the vertices of the dependency tree. $D_{min}$ and $D_{max}$, are the minimum and the maximum value of $D$ in a random linear arrangement. 
A uniformly random linear arrangement is one whose probability is $1/n!$. The expected value of $D$ in a uniformly random linear arrangement (rla) of a given graph is \citep{Ferrer2004b,Ferrer2014f} 
\begin{equation}
D_{rla} = \frac{n^2 - 1}{3}.
\label{expected_sum_of_edge_lengths_equation}
\end{equation}

\subsection{A simple binomial test}

\label{binomial_test_subsection}

Suppose that $f(D > D_{rla})$ is the number of sentences where $D$ exceeds $D_{rla}$. 
We wish to test if $f(D > D_{rla})$ is larger than expected by chance. If that is the case then we conclude that there is evidence of anti-DDm. 

We define {\bf Null hypothesis 1} as a two-fold null model on sentences of length $n$ from an individual treebank:
\begin{enumerate}
\item
The syntactic dependency trees are the same as in the original treebank.
\item
Although the tree structure of every sentence is the same as in the original dataset, vertices are ordered according to a uniformly random linear arrangement. Linear arrangements are independent.
\end{enumerate}
The null model can be applied to a subset of the trees of length $n$, e.g., all star trees of $n$ vertices.

Next subsections analyze the case of sentences of three words and four words ($n=3$ and $n=4$), presenting derivations of binomial distributions from {\bf Null hypothesis 1} or a variant that can help one to find evidence of anti-DDm.  

\subsubsection{$n=3$}

\begin{figure}
\begin{center}
\includegraphics[scale = 0.8]{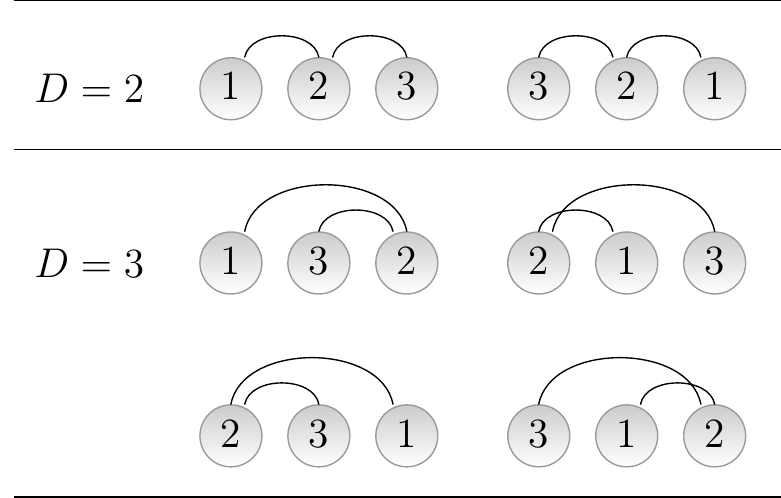}
\end{center}
\caption[All the linear arrangements of a tree with $n=3$ classified by $D$ (the sum of edge distances). For each value of $D$, the arrangements are sorted in ascending lexicographic order according to the sequence defined by vertex labels.]{\label{linear_arrangements_tree_3_vertices_figure} All the linear arrangements of a tree with $n=3$ classified by $D$ (the sum of edge distances). For each value of $D$, the arrangements are sorted in ascending lexicographic order according to the sequence defined by vertex labels. }
\end{figure}

When $n = 3$, $D = 2$ when the hub is put at the center and $D = 3$ when it is not (Figure \ref{linear_arrangements_tree_3_vertices_figure}). Obviously, $D_{min} = 2$ and $D_{max} = 3$. The probability that $D = 2$ in a uniformly random linear arrangement is $p(D=2)=1/3$. To see it, notice that only two orderings have the hub at the center (the non-hub vertex that is put first gives the two orderings; Figure \ref{linear_arrangements_tree_3_vertices_figure}) and that there are $3!$ possible orderings. 
Thus, $p(D=2) = 2/3! = 1/3$.  Then $p(D=3) = 1 - 1/3 = 2/3$.

When $n=3$, Eq. \ref{expected_sum_of_edge_lengths_equation} gives $D_{rla} = 8/3$. Thus, satisfying $D > D_{rla}$ in a sentence is equivalent to satisfying $D = D_{max}$.  
We define an indicator variable $I[D = D_{max}]$ such that 
\begin{equation*}
I[D = D_{max}] = 
   \left\{
   \begin{array}{ll} 
     1 & \mbox{~if~} D = D_{max} \\
     0 & \mbox{~otherwise}
   \end{array}
   \right.
\end{equation*} 
in a dependency tree of $n$ vertices.  
Interestingly, $I[D = D_{max}]$ follows a Bernoulli distribution because
$I[D = D_{max}] = 1$ with probability $2/3$ and 
$I[D = D_{max}] = 0$ with probability $1/3$.

Suppose that $f(D = D_{max})$ is the number of syntactic dependency trees where $D = D_{max}$ and $f(n = 3)$ is the number of syntactic dependency trees where $n = 3$.
Under {\bf Null hypothesis 1}, it turns out that $f(D = D_{max})$ follows a binomial distribution with parameters $f(n = 3)$ and $2/3$. Then, one can test if $f(D = D_{max})$ is significantly large, in favour of anti-DDm, with a binomial test \citep{Conover1999a}.

We will check the probabilities above applying the general definition of 
$D_{rla}$, namely
\begin{equation}
D_{rla} = \sum_{D'=D_{min}}^{D_{max}}p(D = D') D'. 
\label{general_expected_sum_of_edge_lengths_equation}
\end{equation}
When $n=3$, the general formula gives
\begin{eqnarray*}
D_{rla} & = & 2 p(D=2) + 3 p(D=3) \\
                 & = & 2 \frac{1}{3} + 3 \frac{2}{3} \\
                 & = & 8/3
\end{eqnarray*}
as expected by Eq. \ref{expected_sum_of_edge_lengths_equation}.

\subsubsection{$n=4$}
 
When $n=3$, there is only one possible unlabelled tree, that is both a linear tree and a star tree (Figure \ref{labelled_unlabelled_figure}). 
When $n=4$, there are only two possible unlabelled trees: a star tree and a linear tree (Figure \ref{labelled_unlabelled_figure}). We will investigate each kind of tree separately.


\begin{figure}
\begin{center}
\includegraphics[scale = 0.8]{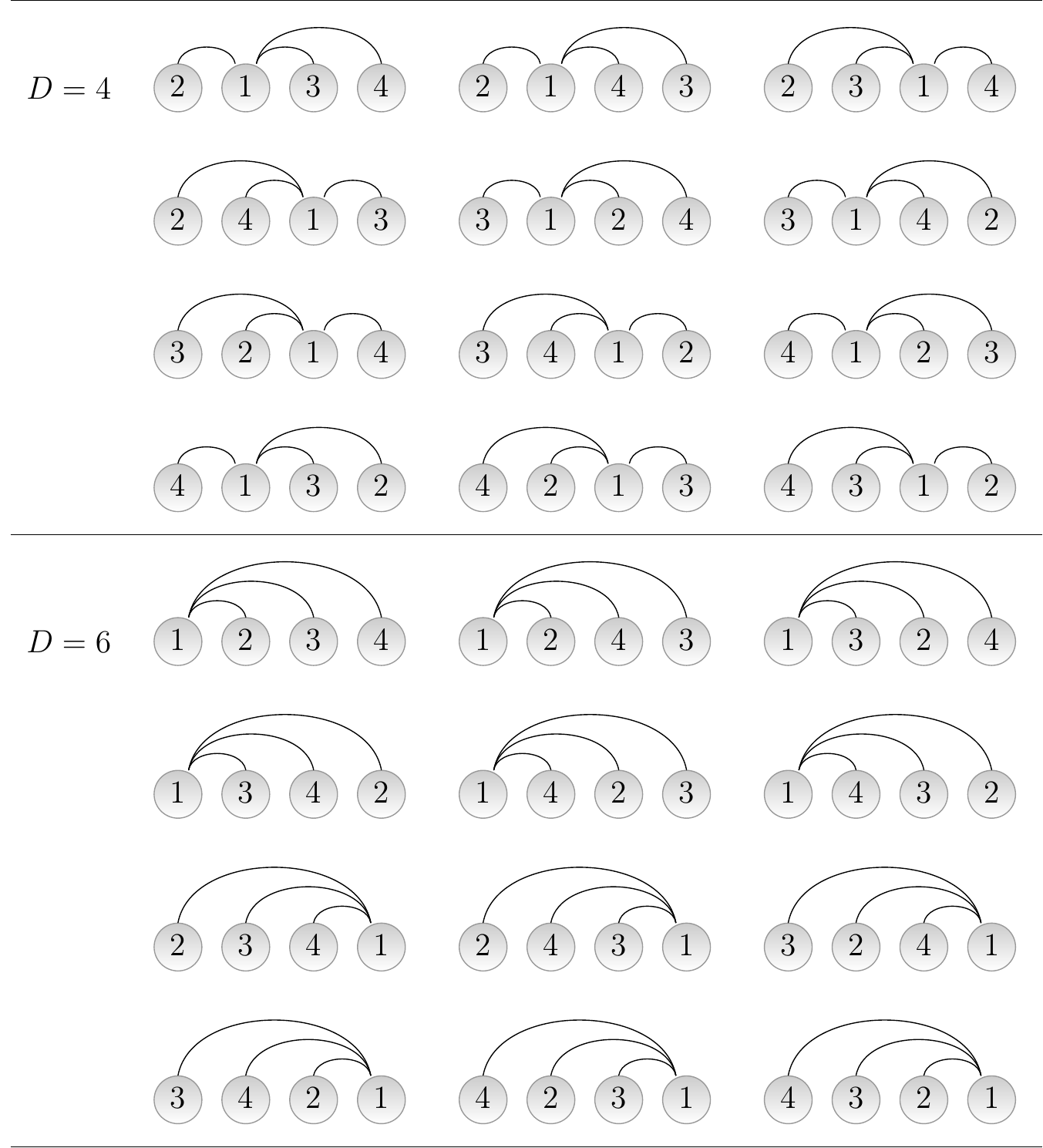}
\end{center}
\caption[All the linear arrangements of a star tree with $n=4$ classified by $D$ (the sum of edge distances). The hub is labelled with $1$. For each value of $D$, the arrangements are sorted in ascending lexicographic order according to the sequence defined by vertex labels.
]{\label{linear_arrangements_star_4_vertices_figure} All the linear arrangements of a star tree with $n=4$ classified by $D$ (the sum of edge distances). The hub is labelled with $1$. For each value of $D$, the arrangements are sorted in ascending lexicographic order according to the sequence defined by vertex labels. }
\end{figure}

In a star tree with $n=4$, there are only two possible values of $D$: $D=4$, when the hub is placed in one of the two central positions, and $D = 6$, when the hub is placed at one of the two ends (Fig. \ref{linear_arrangements_star_4_vertices_figure}). Obviously, $D_{min} = 4$ and $D_{max} = 6$.  
$p_s(D=4)$, the probability that $D = 4$ in a uniformly random linear arrangement of a star tree, is $1/2$. To see it, notice that there are six orderings where the hub is placed in the 1st central position, i.e., the 2nd position (permuting the positions of the non-hub vertices gives 6 configurations). There are six more orderings where the hub is placed in the 2nd central position, i.e., the 3rd position. Thus, $p_s(D=4) = (6+6)/4! = 1/2$.  Then $p_s(D=6) = 1 - 1/2 = 1/2$.

When $n=4$, Eq. \ref{expected_sum_of_edge_lengths_equation} gives $D_{rla} = 5$. Thus, $p_s(D > D_{rla})$, the probability that $D$ exceeds $D_{rla}$ in a random linear arrangement of a star tree with $n= 4$, matches $p_s(D = D_{max})$. Applying the same arguments for the case $n=3$, we obtain that, under {\bf Null hypothesis 1},  
$f(D = D_{max})$ follows a binomial distribution with parameters $f_s(n = 4)$ and $1/2$, where $f_s(n = 4)$ is the number of star trees of 4 vertices. 

We will check the probabilities above applying the general definition of 
$D_{rla}$ in Eq. \ref{general_expected_sum_of_edge_lengths_equation}.
For a star tree with $n=4$, the general formula gives
\begin{eqnarray*}
D_{rla} & = & 4 p_s(D=4) + 6 p_s(D=6) \\
             & = & 4 \frac{1}{2} + 6\frac{1}{2} \\
             & = & 5
\end{eqnarray*}
as expected by Eq. \ref{expected_sum_of_edge_lengths_equation}.


\begin{figure}
\begin{center}
\includegraphics[scale = 0.8]{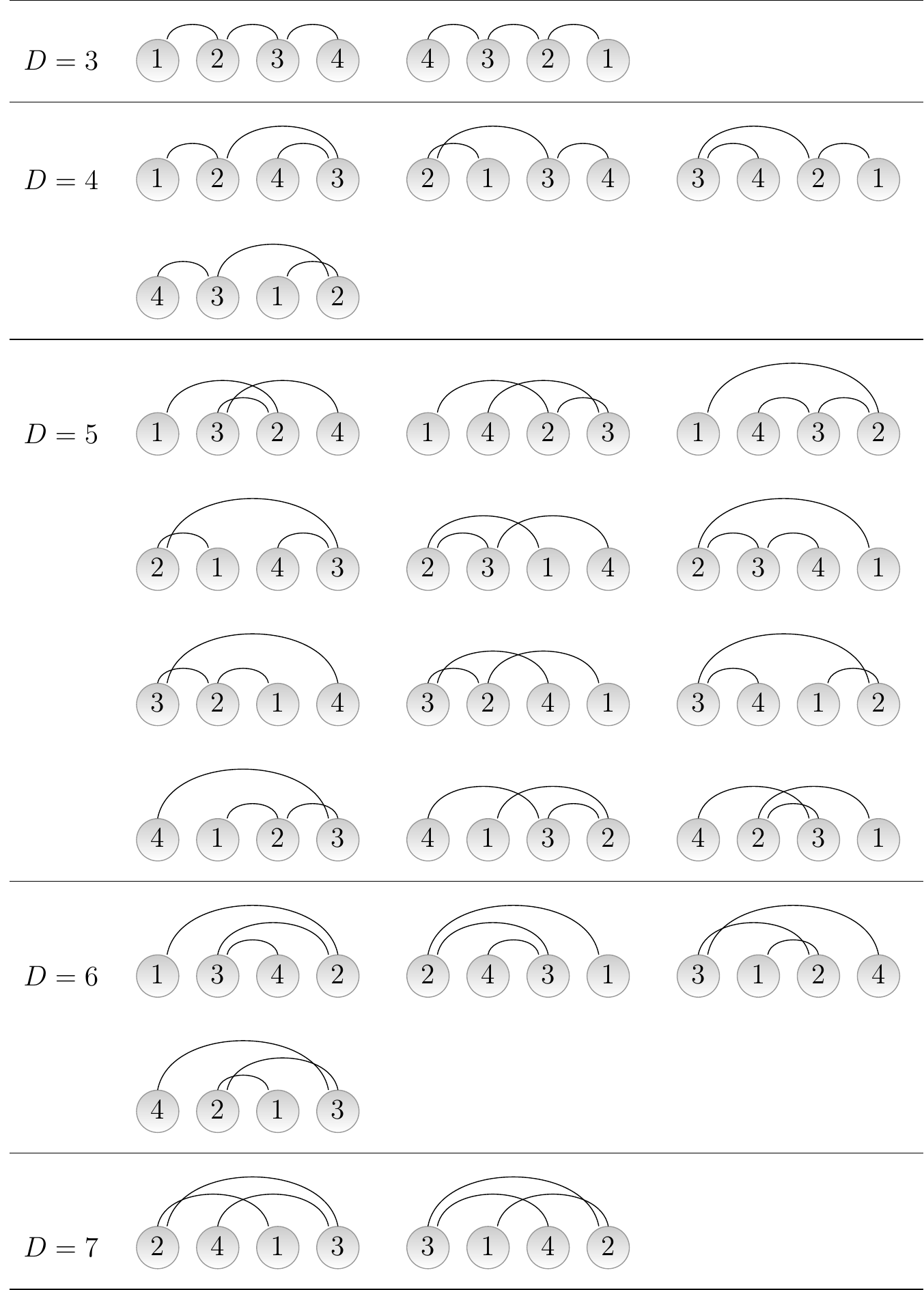}
\caption[All the linear arrangements of a linear tree with $n=4$ classified by $D$ (the sum of edge distances). For each value of $D$, the arrangements are sorted in ascending lexicographic order according to the sequence defined by vertex labels.
]{\label{linear_arrangements_tree_4_vertices_figure} All the linear arrangements of a linear tree with $n=4$ classified by $D$ (the sum of edge distances).  For each value of $D$, the arrangements are sorted in ascending lexicographic order according to the sequence defined by vertex labels. }
\end{center}
\end{figure}

In a linear tree with $n=4$, there are only five possible values of $D$: 3, 4, 5, 6 and 7. Fig. \ref{linear_arrangements_tree_4_vertices_figure} shows all the permutations giving each of the values of $D$ for a total of $4! =24$. Therefore, the probabilities of the values of $D$ in a linear tree are 
\begin{itemize}
\item
$p_l(D=3) = p_l(D=7) = 2/4! = 1/12$.
\item
$p_l(D=4) = p_l(D=6) = 4/4! = 1/6$.
\item
$p_l(D=5) = 12/4! = 1/2$.
\end{itemize}
Applying the probabilities above to the general definition of 
$D_{rla}$ in Eq. \ref{general_expected_sum_of_edge_lengths_equation} one obtains
\begin{eqnarray*}
D_{rla} & = & (3+7)\frac{1}{12} + (4+6)\frac{1}{6} + 5 \frac{1}{2} \\
                 & = & 5
\end{eqnarray*}
as expected by Eq. \ref{expected_sum_of_edge_lengths_equation}.  
Thus, the probability that $D$ exceeds $D_{rla}$ in a random linear arrangement of a linear tree with $n=4$ is
\begin{equation*}
p_l(D > D_{rla}) =  p_l(D \in \{6,7\}).
\end{equation*}
Applying the same arguments for the case $n=3$ or $n=4$ for star trees, we obtain that, under {\bf Null hypothesis 1},  
$f_l(D > D_{max})$, the number of linear arrangements of linear trees of 4 vertices, follows a binomial distribution with parameters $f_l(n = 4)$ and $1/4$, where $f_l(n = 4)$ is the number of linear trees of 4 vertices and $1/4 = p(D = 6) + p(D = 7)$.


Above, we have distinguished linear trees from star trees. Now we consider an arbitrary tree, which leads to  {\bf Null hypothesis 2}, an additional three-fold null model (differences with respect to {\bf Null hypothesis 1} are marked in boldface):
\begin{enumerate}
\item
{\bf The number of sentences of the target length is the same as in the original treebank.}
\item
{\bf The tree structure of every sentence is chosen from a given statistical ensemble. Trees are independent. }
\item
The vertices of each sentence are ordered according to a uniformly random linear arrangement. Linear arrangements are independent.
\end{enumerate}
{\bf Null hypothesis 2} becomes testable when a statistical ensemble is chosen.

We will consider three statistical ensembles: real trees, uniformly random labelled trees, i.e. all labelled trees with the same $n$ are equally likely, and uniformly random unlabelled trees, i.e. all unlabelled trees with the same $n$ are equally likely (Figure \ref{labelled_unlabelled_figure}). 
The choice of uniformity for the two last ensembles can be justified based on the maximum entropy principle without constraints \citep{Kesavan2009a}.

Under {\bf Null hypothesis 2}, $f(D > D_{rla})$, the number of linear arrangements of arbitrary trees of 4 vertices where $D$ exceeds $D_{rla}$, follows a binomial distribution with parameters $f(n = 4)$ and $p(D > D_{rla})$. 
Suppose that $p_s$ is the probability that a tree of 4 vertices is a star tree and $p_l$ the same for a linear tree.  
As $p_s + p_l = 1$, $p(D > D_{rla})$ can be derived noting that
\begin{eqnarray}
p(D > D_{rla}) & = & p_s(D > D_{rla}) p_s + p_l(D > D_{rla}) p_l \nonumber \\
                        & = & p_s(D > D_{rla}) p_s + p_l(D > D_{rla}) (1- p_s) \nonumber  \\
                        & = & [p_s(D > D_{rla}) - p_l(D > D_{rla})] p_s + p_l(D > D_{rla}). \label{probability_equation}  
\end{eqnarray}
Recalling that $p_s(D > D_{rla}) = 1/2$ and $p_l(D > D_{rla}) = 1/4$, Eq. \ref{probability_equation} becomes
\begin{equation}
p(D > D_{rla})  = \frac{1}{4}(p_s  + 1). \label{refined_probability_equation}  
\end{equation}

$p_s$ is determined by the statistical ensemble for which {\bf Null hypothesis 2} is instantiated.
When considering the statistical ensemble of real trees, $p_s$ is the proportion of star trees (with $n=4$) in a treebank.
Knowing that there are only two possible unlabelled trees with $n=4$, star trees and linear trees (Figure \ref{labelled_unlabelled_figure}), we obtain $p_s = 1/2$ for uniformly random labelled trees.
Knowing that there are 16 labelled trees with $n=4$, of which $4$ are star trees (Figure \ref{labelled_unlabelled_figure}), we obtain $p_s = 4/16 = 1/4$ for uniformly random labelled trees. Therefore, Eq. \ref{refined_probability_equation} gives
$p(D > D_{rla}) = 3/8$ for uniformly random unlabelled trees and $p(D > D_{rla}) = 5/16$ for uniformly random labelled trees. 
The fact that $3/8 =0.375 > 5/16 = 0.3125$ allows one to predict that anti-DDm should surface more clearly with uniformly random labelled trees than with uniformly random unlabelled trees.

To sum up, the case of arbitrary trees with $n = 4$ will be investigated with the help of various binomial tests, 
based on the fact that $f(D > D_{rla})$ is binomially distributed with parameters $f(n = 4)$ and $p(D > D_{rla})$. The second parameter depends on the kind of random tree. 

\subsubsection{Simple binomial tests for DDm}

\label{binomial_tests_dependency_length_minimization_subsection}

Applying the same methodology, it is possible do derive binomial tests for the case of DDm. The distribution under the null model turns out to be the same as that of anti-DDm by symmetry except for $n=3$. 
When $n = 3$, $f(D < D_{rla})$ follows a binomial distribution with parameters $f(n=3)$ and $1/3$.
When $n = 4$, 
\begin{itemize}
\item
$f_s(D < D_{rla})$ follows a binomial distribution with parameters $f_s(n=4)$ and $1/2$.
\item
$f_l(D < D_{rla})$ follows a binomial distribution with parameters $f_l(n=4)$ and $1/4$.
\item
$f(D < D_{rla})$ follows a binomial distribution with parameters $f(n=4)$ and probability $(p_s + 1)/4$, that becomes 
$5/16$ in uniformly random labelled trees and $3/8$ in uniformly random unlabelled trees. This is easy to see noting that Eq. \ref{probability_equation} gives
\begin{equation}
p(D < D_{rla}) = [p_s(D < D_{rla}) - p_l(D < D_{rla})] p_s + p_l(D < D_{rla}) 
\label{symmetric_probability_equation}  
\end{equation}
by symmetry. 
\end{itemize} 

\subsubsection{Minimum sample size}

Here we aim to investigate when the sample size $m$, namely the number of sentences involved in a certain binomial test, is too small to allow one to reject the null hypothesis.
Suppose a random variable $f$ that follows a binomial distribution with parameters $m$ and $p$. In our binomial tests, the p-value is the probability that $f$ equals or exceeds a certain value $g$, i.e.   
\begin{equation*}
p\mbox{-value} = \sum_{f = g}^m {m \choose f} p^f (1-p)^{m-f}.
\end{equation*}
As all the summands are positive, the smallest p-value is obtained when $g = m$ and then the p-value is $p^m$. A necessary condition for significance is then
\begin{equation*} 
p^m \leq \alpha.
\end{equation*}
Taking logarithms on both sides of the inequality (and noting this will change the sign of left and the right hand side because $0 < p, \alpha \leq 1$), one obtains
\begin{equation}
m\geq m^*
\label{minimum_sample_size_equation}
\end{equation} 
with
\begin{equation*}
m^* = \left \lceil \frac{\log \alpha}{\log p} \right \rceil.
\end{equation*}
 
Table \ref{minimum_sample_size_table} shows the value of $m^*$ for the different tests except for the case of real trees with $n=4$, where $p$ depends on the proportion of star trees. When Eq. \ref{minimum_sample_size_equation} is not satisfied, the binomial tests suffer from undersampling (Eq. \ref{minimum_sample_size_equation} provides a necessary but not sufficient condition for sufficient sampling).

\begin{table}
\caption[Minimum sample size for each binomial test.]{\label{minimum_sample_size_table} $m^*$ the minimum sample size to achieve significance for each binomial test assuming a significance level $	\alpha = 0.05$. }
\begin{center}
\begin{tabular}{llll}
$n$ & Kind & $p$ & $m^*$ \\
\hline
3 & DLM & $1/3$ & 3 \\
  & anti-DLM & $2/3$ & 8 \\
4 & unlabelled & $3/8$ & 4 \\
  & labelled & $5/16$ & 3  \\
  & star & $1/2$ & 5 \\
  & linear & $1/4$ & 3 \\
\end{tabular}
\end{center}
\end{table}

\subsubsection{Summary}

\label{summary_subsection}

We have shown above that we can test DDm or dependency distance maximization (anti-DDm) with the help of one-tailed binomial tests \citep{Conover1999a}. 
When $n = 3$, $f(D < D_{rla})$ follows a binomial distribution with parameters $f(n=3)$ and $1/3$ whereas $f(D > D_{rla})$ follows a binomial distribution with parameters $f(n=3)$ and $2/3$.
When $n = 4$, 
\begin{itemize}
\item
$f_s(D < D_{rla})$ and $f_s(D > D_{rla})$ follow a binomial distribution with parameters $f_s(n=4)$ and $1/2$.
\item
$f_l(D < D_{rla})$ and $f_l(D > D_{rla})$ follow a binomial distribution with parameters $f_l(n=4)$ and $1/4$.
\item
$f(D < D_{rla})$ and $f(D > D_{rla})$ follow a binomial distribution with parameters $f(n=4)$ and probability $(p_s + 1)/4$, that becomes 
$5/16$ in uniformly random labelled trees and $3/8$ in uniformly random unlabelled trees. 
\end{itemize} 

Given a treebank from a certain language, we consider six levels of application of the binomial test:
\begin{itemize}
\item
All with $n=3$, i.e. any tree with $n = 3$. 
\item 
All with $n = 4$, i.e. any tree with $n = 4$ using real trees for reference ($p_s$ is borrowed from real trees with $n = 4$).
\item
Unlabelled, i.e. any tree with $n = 4$ using random unlabelled trees with $n = 4$ for reference ($p_s$ is borrowed from random unlabelled trees with $n = 4$).
\item
Labelled, i.e., any tree with $n = 4$ using random labelled trees with $n = 4$ for reference ($p_s$ is borrowed from random labelled trees with $n = 4$).
\item
Star trees with $n = 4$.
\item
Linear trees with $n = 4$.  
\end{itemize} 

The binomial tests were carried out with the function \texttt{binom.test} from the R programming language \citep{R2018}.

  
\subsection{The risks of disallowing edge crossings} 

Here we will consider the effect of disallowing crossings for each of the six levels of application of the binomial tests. 
The ban does not have any impact for $n = 3$ or star trees with $n=4$, because crossings are impossible for star trees \citep{Ferrer2013b}.
As for star trees with $n=4$,  Fig. \ref{linear_arrangements_tree_4_vertices_figure} allows one to see that 8 out of 4! arrangements contain crossings: 6 permutations with crossings with $D=5$ and 2 permutations with crossings with $D = 7$.  Then, the number of relevant linear arrangements drops from $4!$ to $4! - 8 = 16$ non-crossing linear arrangements and $D_{rla}$, the expected value of $D$ in non-crossing linear arrangements, can be calculated as the average value of $D$ over these arrangements based on Fig. \ref{linear_arrangements_tree_4_vertices_figure} as 
\begin{eqnarray*}
D_{rla} & = & \frac{1}{16} \left( 2 \cdot 3 + 4 \cdot 4+ 6 \cdot 5 + 4 \cdot 6 + 0 \cdot 7 \right). \\
           & = & \frac{19}{4} = 4.75. \\
\end{eqnarray*}
Accordingly,
\begin{eqnarray*}
p_l(D > D_{rla}) & = & \frac{6 + 4}{16} \\
                 & = & \frac{5}{8}
\end{eqnarray*}
and 
\begin{eqnarray*}
p_l(D < D_{rla}) & = & \frac{2+4}{16} \\
                 & = & \frac{3}{8}.
\end{eqnarray*}
When banning crossings, 
\begin{itemize}
\item
$p_l(D > D_{rla})$ grows from $1/4$ to $5/8$, implying that the test of anti-DDm for linear trees is more likely to make type II errors.
\item 
$p_l(D < D_{rla})$ grows from $1/4$ to $3/8$, implying that the test of DDm for linear trees is also more likely to make type II errors. 
\end{itemize}
At the level of $n=4$, the application of $p_s(D > D_{rla}) = 1/2$ and $p_l(D > D_{rla}) = 5/8$  
to Eq. \ref{probability_equation} gives that the value of $p(D > D_{rla})$ in non-crossing (nc) configurations is
\begin{equation}
p_{nc}(D > D_{rla})  = \frac{1}{8}(5 - p_s). \label{another_refined_probability_no_crossings_equation}  
\end{equation}
It is easy to see that $p_{nc}(D > D_{rla}) \geq p(D > D_{rla})$ and then the binomial tests of anti-DDm when crossings are banned are conservative also at the level of all trees with $n = 4$, labelled trees and unlabelled trees. 

A similar conclusion is reached for the tests of DDm. 
The application of $p_s(D < D_{rla}) = 1/2$ and $p_l(D < D_{rla}) = 3/8$  
to Eq. \ref{symmetric_probability_equation} gives 
\begin{equation}
p_{nc}(D < D_{rla})  = \frac{1}{8}(p_s  + 3). \label{refined_probability_no_crossings_equation}  
\end{equation}
It is easy to see that $p_{nc}(D < D_{rla}) \geq p(D < D_{rla})$ and then the binomial tests of DDm when crossings are banned are conservative also at the level of all trees, labelled trees and unlabelled trees with $n =4$. We conclude that banning crossings precludes the detection of anti-DDm and also DDm when $n=4$, consistent with previous arguments for the case of DDm in a general context \citep{Ferrer2016a}.

\subsection{Additional methods on top of the binomial test}
 
For each level of analysis, we apply a binomial test to check anti-DDm and another to check DDm.
To control for multiple comparisons within each level and target (anti-DDm or DDm), we apply a Holm correction, that does not assume independence between $p$-values \citep{Goeman2014a}. This point is crucial in our case because languages in our sample are not independent, a well-known problem since Galton \citep{Naroll1965a}.
See \citep{Goeman2014a} for a detailed analysis of the minimal assumptions of the correction and how to calculate it.   

Notice that we are applying the correction globally, for all languages available for a certain level of analysis, and then checking if the null hypothesis is rejected in different families to fight against Galton's problem. To control for the relatedness of languages, a simple, although conservative test is to run the analysis within each language family \citep{Roberts2013b}. Accordingly, we may also apply the correction within each family, but this would imply a less effective control for multiple comparisons. This is easy to see mathematically from the standpoint of the Bonferroni correction, the precursors of Holm's correction \citep{Goeman2014a}. Bonferroni's original correction consists of multiplying the p-value by $\lambda$, the number of languages. 
As $\lambda$ is always smaller within a family, that means that applying the correction within family reduces the penalty for multiple comparisons.